\documentclass[10pt, a4paper]{article}

\usepackage[utf8]{inputenc}
           
\usepackage{lrec}
\usepackage{multibib}
\newcites{languageresource}{Language Resources}
\usepackage{graphicx}
\usepackage{tabularx}
\usepackage{soul}
\usepackage{multicol}

\usepackage{epstopdf}

\usepackage{hyperref}
\usepackage{xstring}

\usepackage{epsfig}

\title{Czech Text Document Corpus v 2.0}

\name{Pavel Kr\'al$^{1,2}$, Ladislav Lenc$^{1,2}$}
\address{
\begin{tabular}{cc}
$^1$Dept. of Computer Science \& Engineering & $^2$New Technologies for the Information Society\\
Faculty of Applied Sciences & Faculty of Applied Sciences\\
University of West Bohemia & University of West Bohemia \\
Plze\v{n}, Czech Republic & Plze\v{n}, Czech Republic\\
& \\
& \\
\end{tabular}
}

\abstract{
This paper introduces ``Czech Text Document Corpus v 2.0'', a collection of text documents for automatic document classification in Czech language. 
It is composed of the text documents provided by the Czech News Agency and is freely available for research purposes at~\url{http://ctdc.kiv.zcu.cz/}.
This corpus was created in order to facilitate a~straightforward comparison of the document classification approaches on Czech data.
It is particularly dedicated to evaluation of multi-label document classification approaches, because one document is usually labelled with more than one label.
Besides the information about the document classes, the corpus is also annotated at the morphological layer.
This paper further shows the results of selected state-of-the-art methods on this corpus to offer the possibility of an easy comparison with these approaches.
\\ \newline  }

\begin{document}

\maketitleabstract

\section{Introduction}
Automatic classification (or categorization) of text documents is very important for information organization and storage
because of the significant increase of the amount of electronic documents and the rapid growth of the Internet.

Many efficient approaches have been proposed. They are usually based on supervised machine learning.
The documents are projected into the so-called vector space model, basically using the words as features for various classification algorithms.
The approaches differ in the used methods, however the common point is that all of them need an annotated document corpus to train the parameters.

A sufficient number of the corpora in several languages, particularly in English, is freely available.
However, to the best of our knowledge, the Czech one is missing.

The main goal of this paper consists in presenting a corpus of Czech text documents.
It is composed of real newspaper articles provided by the Czech News Agency (\v{C}TK)\footnote{http://www.ctk.eu/} and is available for research purposes for free.
It is created for a~straightforward comparison of the document classification approaches in Czech.
One document is usually labelled with more than one label, therefore this corpus is usually used for evaluation of multi-label document classification.
Besides the information about the document classes, the corpus is also morphologically annotated.
The morphological annotation has been done fully automatically.

Another research contribution of this paper represents the reported results of selected state-of-the-art methods on this corpus to offer the possibility of an easy comparison with these approaches.

The paper structure is as follows.
The following section presents other text corpora for document classification freely available for research purposes.
Section~\ref{sec:corpus} details our corpus. 
Section~\ref{sec:xp} presents the results of the selected state-of-the art methods on this dataset.
The last section concludes the paper.

\section{Other Text Corpora}
Some important existing text classification corpora in several languages are described below.

\subsection{Reuters-21578} 
Reuters-21578\footnote{http://www.daviddlewis.com/resources/testcollections/reuters21578/} corpus is a collection of 21,578 documents.
The training part is composed of 7769 documents, while 3019 documents are reserved for testing.
The number of possible categories is 90 and the average label/document number is 1.23. 
This dataset is the most frequently used benchmark for English.


\subsection{RCV1-V2}
RCV1-V2\footnote{https://archive.ics.uci.edu/ml/datasets/Reuters+RCV1+RCV2+Multilingual,+Multiview+Text+Categorization+Test+collection}~\cite{lewis2004rcv1} is another text classification test collection which is freely available for research purposes.
It contains about 800,000 manually categorized newswire English stories from Reuters, Ltd.
RCV1 contains English documents, while RCV2 is composed of text documents in French, German, Italian, Spanish and others.
This dataset is also widely used as a benchmarking corpus for English and the languages mentioned above.

\subsection{Other Corpora}
For other corpora dedicated for text categorization, you can visit for instance~\url{http://mulan.sourceforge.net/datasets-mlc.html}.

\section{Corpus Description}
\label{sec:corpus}
\subsection{General Information}
The main part (for training and testing) of the Czech Text Document Corpus v 2.0 is composed of 11,955 real newspaper articles provided by the Czech News Agency.
We provide also a development set which is intended to be used for tuning of the hyper-parameters of the created models.
This set contains 2735 additional articles.

The documents belong to different categories (classes) such as weather, politics, sport, culture, etc.
Each document is associated with one or more labels (classes).
It is thus beneficial to use it for multi-label document classification scenarios.
The multi-label classification task is considerably more important than the single-label classification because it usually corresponds better to the needs of the current applications.

The total category number is 60\footnote{This list is reported in Table~\ref{tab:list} in Section~\ref{sec:ap}.} out of which 37 most frequent ones are used for classification.
The reason of this reduction is to keep only the classes with the sufficient number of occurrences to train the models.
The corpus was annotated by professional journalists from the Czech News Agency.
All documents are further automatically morphologically annotated using UDPipe tool.

\subsubsection{Statistical Information}
Table~\ref{tabstat} shows the statistical information about the corpus\footnote{Development set is excluded from all analyses reported in this section.}.
It shows for instance that lemmatization decreases the vocabulary size from 150,899 to 82,986 which represents the reduction by 45\%.
Another interesting observation is the distribution of the POS tags in this corpus.

\begin{figure}[!htb]
  \centering
  {\epsfig{file=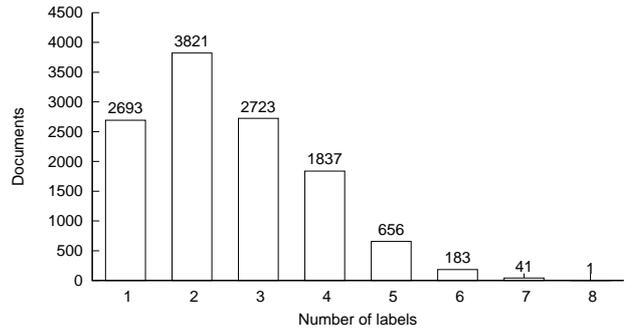,width=8.5cm,angle=0}}
   \caption{Distribution of documents depending on the number of labels}
  \label{fig:doc_labels}
\end{figure}

Figure~\ref{fig:doc_labels} illustrates the distribution of the documents depending on the number of labels.
It shows that the maximal number of categories associated with one document is eight, the majority of documents has two categories and the average label number is 2.55.

\begin{figure}[!htb]
  \centering
  {\epsfig{file=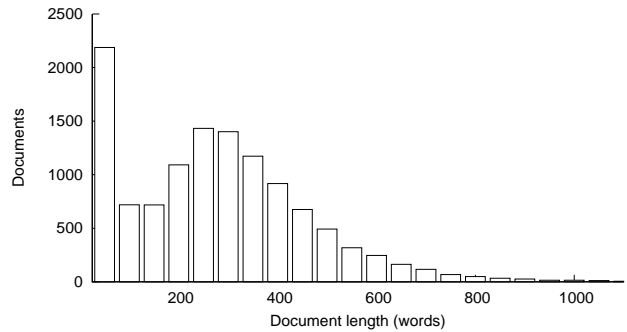,width=8.5cm,angle=0}}
   \caption{Distribution of the document lengths}
  \label{fig:doc_lengths}
\end{figure}

Figure~\ref{fig:doc_lengths} shows the distribution of the document lengths (in word tokens).
The documents are relatively long and the longest documents are composed of more than 7000 word tokens.
Another interesting information is that the most documents (about 2000) contain at most 50 words.
The average document size is 277 words.

\subsubsection{Download}
This dataset is licensed under the Attribution-NonCommercial-ShareAlike 3.0 Unported License\footnote{http://creativecommons.org/licenses/by-nc-sa/3.0/}. 
Therefore it is freely available for research purposes, however any commercial use is strictly excluded.
This corpus is possible to download at~\url{http://ctdc.kiv.zcu.cz/}.

\begin{table}[h!]
\centering
\begin{tabular}{ll|ll}
Unit name & Number & Unit name & Number \\ \hline
 Document  & 11,955 & Word & 3,505,965\\  
 Category & 60 & Unique word  & 150,899\\
 Cat. classif. & 37 & Unique lemma & 82,986 \\
 \hline
Noun	  & 894,951 & Punct	& 553,099 \\
Adjective & 369,172 & Adposition	 & 340,785 \\
Verb	  & 287,253 & Numeral &	265,430 \\
Pronoun   & 258,988 & Adverb &	144,791 \\
Coord. conj.   & 100,611 & Determiner & 84,681 \\
Pronoun   & 74,340         & Aux. verb & 70,810 \\
Subord. conj.  & 41,487 & Particle	& 12,383 \\
Symbol &	2420 & Interjection	& 142\\
Other &	4126  & & \\
\end{tabular}                                                                         
\caption{\label{tabstat}Corpus statistical information}
\end{table}

\subsection{Morphological Annotation}
As already mentioned, we used UDPipe tool~\cite{udpipe:2017}\footnote {http://ufal.mff.cuni.cz/udpipe} for automatic morphological analysis of the corpus.
This tool provides en efficient pipeline for sentence segmentation, tokenization, POS tagging, lemmatization and dependency parsing.
It also contains models for 50 languages of universal dependencies (UD) 2.0.
This system can be further used with data in CoNLL-U format\footnote{http://universaldependencies.org/format.html}.

According to the authors, the accuracy of the lemmatizer and of the POS tagger are both about 98\% on the UD version 2 of the Prague Dependency Treebank (PDT) 3.0\footnote{http://ufal.mff.cuni.cz/pdt3.0/}.
The performance of the syntactic parsing is represented by labelled attachment score (LAS) which is about 83\%.
This system uses 17 part-of-speech categories drawn from the revised version of the Google universal POS tags.

\subsection{Technical Details}
Text documents are stored in the individual text files using UTF-8 encoding.
Each filename is composed of the serial number and the list of the categories abbreviations separated by the underscore symbol and the {\it.txt} suffix.
Serial numbers are composed of five digits and the numerical series starts from the value one.

For instance the file {\it 00046\_kul\_nab\_mag.txt} represents the document file number 46 annotated by the categories {\it kul} (culture), {\it nab} (religion) and {\it mag} (magazine selection).
The content of the document, i.e. the word tokens, is stored in one line. The tokens are separated by the space symbols.

Every text document was further automatically mophologically analyzed.
This analysis includes lemmatization, POS tagging and syntactic parsing.
The fully annotated files are stored in {\it .conll} files.
We also provide the lemmatized form, file with suffix {\it .lemma}, and appropriate POS-tags, see {\it .pos} files.
The tokenized version of the documents is also available in {\it .tok} files.

\subsection{Evaluation Protocol}

All following experiments use the five-fold cross validation procedure,
where 20\% of the corpus is reserved for testing and the remaining part for training of the models.
The development set was used to tune the hyper-parameters of the models.

For evaluation of the multi-label document classification, it is used the standard  Precision (P), Recall (R) and F-measure ({\it F$_m$}) metrics~\cite{powers2011evaluation}.
For evaluation of the single-label classification, the authors use the standard accuracy metric.

The confidence interval of the experimental results is~0.6\% at a~confidence level of 0.95~\cite{press1996numerical}.

\section{Experiments}
\label{sec:xp}
The two following sections present the results of the selected classification algorithms on this dataset.
The first section deals with multi-label document classification while the second one describes the classification score of single-label classification task.

\subsection{Multi-label Document Classification}
The first reported approach~\cite{Kral13} uses Bag of Words (BoW) to create the features.
Non-significant words are removed using Part of speech (POS) filtering and for feature selection, the mutual information method is used.
In the original paper, the authors show the results of three classifiers, namely Naive Bayes, Maximum entropy (ME) and Support vector machine with three traditional multi-label classification approaches.
Only the best classification accuracy obtained by ME classifier is reported in this paper.

The second method~\cite{Kral14MICAI} proposes novel unsupervised features using an unsupervised stemmer, latent Dirichlet allocation and semantic spaces (HAL and COALS).
These features are integrated with word features to improve classification results.
Multi-label classification scenario is realized using a set of binary classifiers.
Maximum entropy model is used for classification.

Neural networks are very popular in natural language processing field today and they outperform many state-of-the-art approaches with only very simple preprocessing.
The following approach~\cite{lenc2017deep} uses two different feed-forward neural networks, namely multi-layer perceptron (MLP) and convolutional neural networks (CNN) to achieve new state-of-the art results on this corpus.
The authors use thresholding to realize multi-label document classification task.

The results of the above described approaches are illustrated in Table~\ref{tab:baseline1}.

\begin{table}[!h]
\centering
\begin{tabular}{p{5.0cm}p{0.5cm}p{0.5cm}|p{0.5cm}}
\hline
Method & $P$  & $R$ & $F_m$ \\
\hline
ME~\cite{Kral13}		& - & - & 76.8	\\ 
words+ME~\cite{Kral14MICAI}	&  88.1& 72.7&	79.7\\ 
unsup+ME \cite{Kral14MICAI}	& 89.0&	75.6& 81.7 \\ 
MLP~\cite{lenc2017deep}	& 83.7& 83.6& 83.9 \\
CNN~\cite{lenc2017deep}	& 86.4& 82.8& 84.7 \\
\hline
\end{tabular}
\caption{\label{tab:baseline1}Multi-label document classification results of the different approaches}
\end{table}

\subsection{Single-label Document Classification}
The authors~\cite{hrala13} evaluate five feature selection methods and three classifiers on this corpus.
Lemmatization and POS tagging are used for a precise representation of the Czech documents.
It was demonstrated that POS-tag filtering is very important, while the lemmatization plays only a marginal role for classification.
In this work, only the first document class was considered for classification and the authors consider it as the main document category.
The best classification accuracy was obtained by SVM classifier and is 91.2\%.

\section{Conclusions} 
This paper introduced a novel collection of Czech text documents.
This corpus is composed of real newspaper articles provided by the Czech News Agency and is available for research purposes for free.
It was created to facilitate a~straightforward comparison of the document classification approaches in Czech language.

This corpus is particularly intended to evaluate multi-label document classification approaches,
because one document is usually associated with more than one label.
Besides the information about the document classes, the corpus is automatically annotated at morphological layer.
This paper further shows the results of the selected state-of-the-art algorithms on this corpus to offer the possibility of a straightforward comparison with the future research.

We plan to submit this corpus to be a part of the Language Research Infrastructure of LINDAT/CLARIN project\footnote{https://lindat.mff.cuni.cz/en}.

\section{Acknowledgements}
This work has been supported by the project LO1506 of the Czech Ministry of Education, Youth and Sports.

\section{Language Resource References}
\bibliographystyle{lrec}
\bibliography{paper}

\section{Appendix}
\label{sec:ap}

\begin{table}[!b]
\centering
{\footnotesize
  \begin{tabular}{p{0.5cm}p{3.25cm}|p{3.3cm}}
Abbr.  & Category in Czech &  English translation \\
\hline
aut &  Automobilový průmysl & Automobile industry\\ 
bos &  Bohemika & Czech Rep. from abroad\\
bsk &  Sklářský průmysl & Glass industry\\
bua &  Burzy akciové & Stock exchanges\\
buk &  Burzy komoditní & Commodity exchanges\\
bup &  Burzy peněžní & Currency exchanges\\
bur &  Burzy & Exchanges\\
cen &  - & - \\
che &  Chemický a farmaceutický průmysl & Chemical and pharmaceutical industry \\
den &  Zpravodajské deníky & News schedules \\ 
dpr &  Doprava & Transport \\
dre &  Dřevozpracující průmysl & Woodworking industry \\
efm &  Firmy & Companies \\
ekl &  Životní prostředí & Environment \\
eko &  Ekologie & Ecology\\
ene &  Energie & Energy \\
eur &  Evropská unie - zprávy & European union - news\\
fin &  Finanční služby & Financial services \\
for &  Parlamenty a vlády & Parliaments and governments \\
fot &  Fotbal - zprávy &  Soccer \\ 
hok &  Hokej - zprávy & Ice hockey \\
hut &  Hutnictví & Metallurgy \\
kat &  Neštěstí a katastrofy & Accidents and disasters \\
kul &  Kultura &  Culture \\
mag &  Magazínový výběr  & Magazine selection \\
mak &  Makroekonomika & Macroeconomics \\
med &  Média a reklama & Media and advertising \\
met &  Počasí & Weather \\
mix &  Mix & Mix \\
mot &  Motorismus & Motoring \\ 
nab & Náboženství & Religion \\
obo & Obchod & Trade \\
odb & Práce a odbory & Labour and Trade Unions \\
pit & Telekomunikace a IT & Telecommunications \& IT\\
pla & Plány zpravodajství ČTK & Events news\\
pod & Politika ČR & Czech Republic Politics\\
pol & Politika & Politics \\
prg & Pragensie & Prague issues \\
prm & Lehký průmysl  & Light industry \\
ptr & Potravinářství & Food industry \\ 
reg & Region & Region \\
sko & Školství & Educational system \\
slo & Slovenika & Slovakia from abroad \\
slz & Služby & Services \\
sop & Sociální problematika & Social problems \\
spc & - & - \\
spl & Životní styl & Life style \\
spo & Sportovní zpravodajství & Sports \\
sta & Stavebnictví a reality & Building industries and property \\
str & Strojírenství & Mechanical engineering \\ 
sur & Suroviny & Raw materials \\
tlk & Telekomunikace & Telecommunications \\
tok & Textil & Textile \\
tur & Cestovní ruch & Tourism \\
vat & Věda a technika & Science and technology \\
zah & Zahraniční & Foreign \\
zak & Kriminalita a právo & Criminality and law \\
zbr & Zbraně & Arms \\
zdr & Zdravotnictví & Health service \\
zem & Zemědělství & Agriculture \\
\hline
\end{tabular}
}
\caption{\label{tab:list}List of the categories}
\end{table}


\end{document}